\documentclass[conference]{IEEEtran}
\IEEEoverridecommandlockouts

\usepackage{cite}
\usepackage{amsmath,amssymb,amsfonts}
\usepackage{algorithmic}
\usepackage{subcaption}
\usepackage{graphicx}
\usepackage{textcomp}
\usepackage{xcolor}
\def\BibTeX{{\rm B\kern-.05em{\sc i\kern-.025em b}\kern-.08em
    T\kern-.1667em\lower.7ex\hbox{E}\kern-.125emX}}

\usepackage{enumitem}
\begin{document}


\title{Leveraging Expert Input for Robust and Explainable AI-Assisted Lung Cancer Detection in Chest X-rays}


\author{\IEEEauthorblockN{Amy Rafferty}
\IEEEauthorblockA{\textit{School of Informatics} \\
\textit{Universiy of Edinburgh}\\
Edinburgh, UK \\
0000-0002-5120-966X}
\and
\IEEEauthorblockN{Rishi Ramaesh}
\IEEEauthorblockA{\textit{NHS Lothian} \\
Edinburgh, UK \\
}
\and
\IEEEauthorblockN{Ajitha Rajan}
\IEEEauthorblockA{\textit{School of Informatics} \\
\textit{University of Edinburgh}\\
Edinburgh, UK \\
0000-0003-3765-3075}
}

\maketitle

\begin{abstract}

Deep learning models show significant potential for advancing AI-assisted medical diagnostics, particularly in detecting lung cancer through medical image modalities such as chest X-rays. However, the black-box nature of these models poses challenges to their interpretability and trustworthiness, limiting their adoption in clinical practice. This study examines both the interpretability and robustness of a high-performing lung cancer detection model based on InceptionV3, utilizing a public dataset of chest X-rays and radiological reports. We evaluate the clinical utility of multiple explainable AI (XAI) techniques, including both post-hoc and ante-hoc approaches, and find that existing methods often fail to provide clinically relevant explanations, displaying inconsistencies and divergence from expert radiologist assessments. To address these limitations, we collaborated with a radiologist to define diagnosis-specific clinical concepts and developed ClinicXAI, an expert-driven approach leveraging the concept bottleneck methodology. ClinicXAI generated clinically meaningful explanations which closely aligned with the practical requirements of clinicians while maintaining high diagnostic accuracy. We also assess the robustness of ClinicXAI in comparison to the original InceptionV3 model by subjecting both to a series of widely utilized adversarial attacks. Our analysis demonstrates that ClinicXAI exhibits significantly greater resilience to adversarial perturbations. These findings underscore the importance of incorporating domain expertise into the design of interpretable and robust AI systems for medical diagnostics, paving the way for more trustworthy and effective AI solutions in healthcare.

\end{abstract}


\section{Introduction}

The deployment of deep learning (DL) models in critical domains such as medical diagnostics is often hindered by their "black-box" nature, which obscures the rationale behind their predictions \cite{b1}. This lack of transparency poses challenges for trust, reliability, and accountability, particularly in high-stakes environments such as the medical domain where errors can cause significant issues regarding patient safety. In recent years, two central concerns have emerged in addressing these challenges: interpretability and robustness. Interpretability is the ability to understand and explain a model's decision-making process, which is essential for clinical adoption, regulatory compliance, and the effective integration of AI-assisted systems in medical workflows \cite{b2} \cite{b3}. Robustness refers to a model’s resilience against adversarial perturbations, data distribution shifts, and noisy inputs, all of which are prevalent in real-world medical applications \cite{b4} \cite{b5}. Adversarial attacks can be categorized into white-box attacks, where the attacker has complete access to the model’s weights, parameters, and architecture, and black-box attacks, where the attacker is limited to observing only the model's inputs and outputs \cite{b6}. Misinterpretations or incorrect classifications can jeopardize patient safety, further underscoring the importance of interpretable and robust models for medical diagnotics \cite{b7}. Addressing these issues is critical to fostering trust in DL systems and ensuring their practical deployment in healthcare.

Explainable artificial intelligence (XAI) techniques have been developed to enhance the interpretability of DL models by providing accessible and understandable explanations for their predictions. XAI methods can be broadly categorized into two classes: post-hoc approaches, which generate explanations after a model is trained, and ante-hoc approaches, which are intrinsically designed to produce interpretable outputs during the prediction process. Despite their promise, the performance of popular XAI techniques often degrades in specialized domains like medical diagnostics, compared to general computer vision tasks \cite{b8} \cite{b9}. For instance, post-hoc methods such as Local Interpretable Model-agnostic Explanations (LIME) \cite{b10} and  Gradient-weighted Class Activation Mapping (Grad-CAM) \cite{b11} frequently fail to highlight critical clinical features in medical images \cite{b12} \cite{b13} \cite{b14} \cite{b15} \cite{b16}, while textual approaches like LLaVA-Med \cite{b17} are prone to omitting pathology-specific details that are vital for diagnosis \cite{b18} \cite{b19}. Contributing factors to these limitations include the poor quality of public medical datasets and inaccuracies in their annotations \cite{b20} \cite{b21}. However, a major limitation, as hypothesized in this work, lies in the absence of expert input during the design of these XAI techniques.

Among ante-hoc XAI methods, Concept Bottleneck Models (CBMs) have become popular for their structured approach to interpretability \cite{b22}. CBMs introduce an interpretable intermediate layer to the classification process by dividing it into two stages: the prediction of high-level concepts from input data, followed by the use of these concepts to determine the final classification. This approach provides full transparency at the concept level, allowing experts to inspect and validate the intermediate representations. In medical diagnostics, for example, these concepts could correspond to specific clinical indications of pathologies observed in chest X-rays, aiding clinicians in the diagnostic process. Despite their potential, existing CBMs rely heavily on unsupervised concepts without leveraging domain-specific expertise, which may limit their clinical relevance \cite{b23} \cite{b24}.

In this study, we evaluate the clinical utility of various existing XAI techniques in the domain of lung cancer detection using a large public dataset of chest X-rays, MIMIC-CXR \cite{b25} \cite{b26} \cite{b27}.

We begin by evaluating post-hoc image-based explainability methods LIME \cite{b10}, Shapley Additive Explanations (SHAP) \cite{b28}, and Grad-CAM \cite{b11}, applied to a high-performing 42-layer InceptionV3 classification model \cite{b29}. Despite the high classification performance of this model, our analysis reveals that these methods consistently fail to highlight the clinically significant regions in the chest X-rays to which they are applied. The image-based explainability methods are assessed based on their overlap with each other and their alignment with the medical ground truth on a subset of cancerous chest X-rays from the VinDr-CXR dataset \cite{b27} \cite{b30} \cite{b31}, which includes pathology location annotations. Both overlap metrics are observed to be extremely low.

We then evaluate textual XAI methods, specifically Chest X-ray Large Language and Vision Assistant (CXR-LLaVA) \cite{b32}, a multimodal large language model (LLM) that generates radiological reports for an input chest X-ray using a question-answering framework, and Cross-Modal Conceptualization in Bottleneck Models (XCBs) \cite{b52}, an unsupervised CBM that learns intermediate clinical concepts from chest X-rays and their corresponding reports via a cross-attention mechanism. Both methods are evaluated based on the proportion of ground truth clinical concepts—extracted from the radiological report associated with a chest X-ray in the MIMIC-CXR dataset—that are reflected in their explanations. CXR-LLaVA exhibits a high false-negative rate in lung cancer detection, frequently focusing on unrelated conditions such as pneumothorax or cardiomegaly, likely due to imbalances in its training data \cite{b32}. XCBs demonstrate relatively strong performance by frequently identifying pathology-specific clinical concepts, but also detect less relevant concepts that do not contribute meaningfully to the explanation and were deemed confusing by a radiologist, and exhibit a high rate of false positives. This limitation likely stems from the unsupervised nature of the learned concepts and the absence of expert guidance in their definition.

All XAI methods are further evaluated for their clinical utility, as determined by an expert radiologist who reviewed a subset of explanations generated by each technique and rated them based on their clinical relevance. The results reveal that all image-based methods received consistently low scores across all images in the subset. In contrast, the text-based methods produced more variable results, sometimes generating fully clinically relevant explanations, but were ultimately deemed unreliable as they also frequently generated explanations that were entirely clinically irrelevant.

To address these poor results, we propose ClinicXAI, an expert-driven CBM approach that incorporates radiologist-curated clinical concepts. Using natural language processing (NLP) techniques, these concepts are automatically extracted from radiology reports during model training, ensuring alignment with real-world clinical practices.

ClinicXAI achieves superior classification performance compared to both the standard InceptionV3 model and XCBs. Its clinical relevance is evaluated by comparing the proportion of ground truth clinical concepts captured in its explanations against those of CXR-LLaVA and XCBs, showing a significant improvement. Furthermore, a radiologist's assessment of ClinicXAI-generated explanations highlighted their consistent and near-complete clinical relevance. These results demonstrate a substantial enhancement in both the clinical utility and correctness of the explanations produced by our expert-driven approach, while maintaining high classification performance.

Finally, we evaluate the robustness of ClinicXAI under adversarial conditions using three widely adopted attack methods: Fast Gradient Sign Method (FGSM) \cite{b33}, Projected Gradient Descent (PGD) \cite{b34}, and Simple Black-Box Attack (SimBA) \cite{b35}. FGSM and PGD are white-box attacks that exploit model gradients, while SimBA is a black-box attack that iteratively modifies input perturbations based on output feedback. We find that ClinicXAI exhibits significantly greater robustness to adversarial perturbations compared to the standard InceptionV3 model, and after completing adversarial training using the adversarial images generated by each of these attacks, this robustness is increased even further.

ClinicXAI is not presented as a groundbreaking contribution, but rather as an illustration of the substantial positive influence that integrating expert input can have on enhancing both the interpretability and robustness of AI systems in the medical domain.

\section{Background}

In this section, we provide an overview of adversarial methodologies and XAI techniques utilized in recent studies. We first discuss adversarial attack and defence strategies implemented to assess and enhance model robustness against adversarial perturbations, followed by an examination of the different categories of XAI approaches employed to evaluate the interpretability of deep learning models in the field of medical imaging.

\subsection{Adversarial Attack Strategies}
Adversarial attacks present a significant challenge to deep learning models by introducing subtle, often imperceptible perturbations to input data, which can significantly degrade model performance \cite{b4} \cite{b33}. These attacks are particularly problematic in critical domains like medical diagnostics, where errors in AI-assisted systems can jeopardize patient safety and undermine the reliability of automated diagnostic tools \cite{b7}.

Attacks are generally classified into white-box and black-box categories based on the attacker's knowledge of the model's internal structure. White-box attacks assume full access to the model’s architecture, parameters, and training data, which makes them highly effective \cite{b36}. Techniques in this category include the Fast Gradient Sign Method (FGSM) \cite{b33} and Projected Gradient Descent (PGD) \cite{b34}. FGSM generates adversarial examples by perturbing the input in the direction of the gradient of the loss with respect to the input image, while PGD iteratively maximizes the model’s loss while constraining the magnitude of the perturbation.

In contrast, black-box attacks assume that the attacker has no access to the internal details of the model. These attacks are more representative of real-world scenarios, particularly in medical diagnostics where access to model internals is unlikely \cite{b37}. They are typically based on transfer-based techniques, such as the Iterative Fast Gradient Sign Method (I-FGSM) \cite{b38},which use surrogate models to generate adversarial examples, or score-based methods, such as Simple Black-Box Attack (SimBA) \cite{b35} and Natural Evolution Strategy (NES) \cite{b39}, which rely on repeated queries to the model to estimate gradients.

\subsection{Adversarial Defense Strategies}
In response to the growing threat of adversarial attacks, significant research has focused on developing defense strategies to improve model robustness. Adversarial training \cite{b40} is one of the most widely studied defense techniques - the model is trained on adversarial examples to enhance its ability to recognize and withstand such perturbations. This approach helps improve model resilience during inference by allowing it to generalize better to adversarial inputs.

Other defensive techniques include input transformations \cite{b41}, which modify the input data before processing by the model to reduce the impact of adversarial perturbations. Examples include image denoising or smoothing to remove noise from the input, improving the model's accuracy on perturbed data. Randomization strategies \cite{b42} introduce stochastic elements into the model’s inference process, making it harder for adversaries to predict the model's behavior. By adding variability to input features or model parameters, these strategies reduce the effectiveness of adversarial attacks. Additionally, model ensembles \cite{b43} combine multiple models with diverse architectures to improve robustness. Even if one model is compromised, the others in the ensemble are less likely to be affected, thus enhancing the overall reliability of the system.

\subsection{Post-Hoc Image-Based XAI}
Post-hoc image-based XAI techniques are primarily categorized into perturbation-based and gradient-based methods. Perturbation-based techniques, such as LIME \cite{b10} and SHAP \cite{b28}, generate random masks on the input image to assess the impact of different regions on the model’s classification decision. Gradient-based techniques, such as Grad-CAM \cite{b11} and Layer-Wise Relevance Propagation (LRP) \cite{b44}, compute the importance of each input neuron by performing a forward or backward pass through the deep learning model. These approaches generally output a heatmap of pixel importance scores, which is used as an explanation for a model's classification decision.

Despite their popularity, recent studies have highlighted the instability of post-hoc XAI explanations when applied to deep learning models in medical imaging, such as in lung pathology classification. Explanations generated by these techniques are often inconsistent with one another, with the medical ground truth, and with expert radiologists' opinions \cite{b15} \cite{b16} \cite{b13}. This inconsistency raises concerns about the clinical applicability of these methods, as they may highlight irrelevant regions in medical scans or miss important clinical features, despite high model classification accuracy \cite{b14}.

\subsection{LLM-Based XAI}
The rise of large language models (LLMs) like ChatGPT \cite{b45} and multimodal models such as Large Language and Vision Assistant (LLaVA) \cite{b46} has introduced new possibilities for textual XAI. These models connect vision encoders with LLMs, often in a chatbot format, to generate explanations for input images via Visual Question Answering (VQA) \cite{b47} \cite{b48}. In the medical domain, specialized chatbots like LLaVA-Med \cite{b17} and CXR-LLaVA \cite{b32} have been developed for generating explanations relevant to chest X-ray images. In this work, we examine the effectiveness of CXR-LLaVA, pre-trained on chest X-ray datasets such as MIMIC-CXR, for producing clinically relevant lung cancer explanations.

\subsection{Concept Bottleneck Models (CBMs)}
Concept Bottleneck Models (CBMs) \cite{b22} represent an ante-hoc XAI approach that splits the traditional classification pipeline into two separate stages: one for predicting the presence of a predefined set of concepts and another for predicting the final label based on these concepts. While CBMs offer insight into the decision-making process by providing intermediate concepts, they require a fully annotated training dataset, which is often infeasible and costly in domains like medical imaging.
Several approaches have been proposed to address this limitation. Post-hoc CBMs \cite{b23} apply the CBM framework to existing classifiers without requiring concept annotations during training. Approaches like label-free CBMs \cite{b24} and language model-guided CBMs \cite{b49} use LLMs like GPT-3 \cite{b50} and Contrastive Language-Image Pre-training (CLIP) models \cite{b51} to automatically generate and label concepts from the data. Cross-Modal Conceptualization in Bottleneck Models (XCBs) \cite{b52} extract concepts from radiology reports associated with chest X-rays using unsupervised cross-attention mechanisms.

While these unsupervised methods have shown promise, their clinical relevance remains uncertain due to the lack of expert input utilized in their design. To address this, we propose ClinicXAI, an expert-driven CBM approach. Instead of relying on unsupervised learning, ClinicXAI leverages common NLP techniques to automatically extract radiologist-curated key phrases indicative of pathologies from the radiology reports associated with the chest X-rays in our dataset. This expert-driven process ensures that the extracted concepts have direct diagnostic interpretability and are customizable based on expert input. 

We argue that this approach offers an effective compromise between fully annotated datasets and unsupervised concept learning, requiring only a one-time input of diagnostic phrases by an expert radiologist. Once gathered, these concept lists can be reused across different datasets and medical imaging modalities, minimizing the burden on expert involvement while maintaining clinical relevance. To further facilitate the adoption of ClinicXAI, we include the concepts used in our study in Table~\ref{tab:concepts}, so that others using the method will not need to gather these concepts themselves. 

\begin{table*}[tb]
\caption{Clinical Concept Selection. Original report phrases are clustered to create clinical concepts for each pathology.}
\begin{center}
\begin{tabular}{|c|c|c|}
\hline
\textbf{Label}&\textbf{Clinical Concepts}&\textbf{Original Phrases} \\
\hline
\textbf{Healthy} & Unremarkable & Normal; Unremarkable; Lungs clear$^{\mathrm{b}}$; No evidence; No interval change$^{\mathrm{b}}$; \\
& & No acute cardiopulmonary abnormality$^{\mathrm{b}}$; Normal hilar contours$^{\mathrm{b}}$; No acute process$^{\mathrm{b}}$ \\
\hline
\textbf{Lung Cancer} & Mass & Mass; Cavitary lesion$^{\mathrm{b}}$; Carcinoma; Neoplasm; Tumor/Tumour; Rounded opacity$^{\mathrm{b}}$; Lung cancer; \\
& & Apical opacity; Lump; Triangular opacity; Malignant; Malignancy \\
\cline{2-3}
& Nodule & Nodular densities/density; Nodular opacities/opacity; Nodular opacification; Nodule \\
\cline{2-3}
& Irregular Hilum & Hilar mass; Hilar opacity; Hilus enlarged$^{\mathrm{b}}$, Hilus fullness$^{\mathrm{b}}$, Hilus bulbous$^{\mathrm{b}}$ \\
\cline{2-3}
& Adenopathy & Mediastinal lymphadenopathy; Mediastinal adenopathy; Hilar lymphadenopathy; Hilar adenopathy \\
\cline{2-3}
& Irregular Parenchyma & Pulmonary metastasis; Carcinomatosis; Metastatic disease \\
\hline
\multicolumn{2}{l}{$^{\mathrm{b}}$Type B: Word order may vary.}
\end{tabular}
\label{tab:concepts}
\end{center}
\end{table*}

\section{Materials and Methods}

In this section, we introduce the datasets used in this study: MIMIC-CXR and VinDr-CXR. MIMIC-CXR serves as the primary dataset for XAI evaluation due to its large size and widespread use in the field. It is also utilized for training and evaluating our expert-driven CBM, ClinicXAI. The VinDr-CXR dataset is employed as a supplementary resource for evaluating image-based XAI methods, as it includes a subset of chest X-rays with pathology location annotations (bounding boxes), enabling assessment of the clinical correctness of image-based explanations. We also introduce our clinical concept extraction approach, which is used to evaluate text-based XAI methods, and for the design of ClinicXAI.

\subsection{MIMIC-CXR}
We use the publicly available, anonymized MIMIC-CXR dataset from PhysioNet, which includes chest X-rays and corresponding free-text radiology reports \cite{b25} \cite{b26} \cite{b27}. This dataset contains X-rays captured from various angles (PA, AP, lateral), but we focus specifically on the standard Posterior-Anterior (PA) view to minimize potential confounding factors \cite{b53}. For cancer detection, we select images labelled as either cancerous (Lung Lesion) or healthy (No Finding), under radiologist guidance, resulting in 43,447 image-label pairs, with a class distribution of 41,063 healthy and 2,384 cancerous samples. To address class imbalance, we apply One-Sided Selection undersampling \cite{b54}, reducing the dataset to 19,478 pairs, with a revised distribution of 17,094 healthy and 2,384 cancerous samples. We use stratified 10-fold cross-validation for model training and evaluation, with each fold consisting of a training set of 17,530 image-report pairs and a test set of 1,948 image-report pairs. Images are resized to 512x512 pixels, normalized, and cropped to remove black background.

\subsection{VinDr-CXR}
To assess the clinical utility of post-hoc image-based XAI methods, we also use the PhysioNet VinDr-CXR dataset \cite{b27} \cite{b30} \cite{b31}, which contains chest X-rays annotated with both pathology labels and bounding boxes indicating the locations of pathologies. For the evaluation described in Section~\ref{evalimagexai}, we select all 826 cancerous chest X-rays from this dataset that include pathology location annotations provided by three radiologists.

\subsection{Clinical Concept Extraction}\label{conceptextraction}

In this section, we detail the methodology used to extract clinical concepts from free-text radiological reports. This approach is applied in this study for evaluating text-based XAI techniques and for designing and assessing our expert-driven concept bottleneck model, ClinicXAI.

We asked a consultant radiologist with over 10 years of experience to analyze a randomly selected subset of radiological reports and identify words and phrases indicative of our two target classes: Healthy and Lung Cancer. Single expert labelling is a common practice in radiology AI research \cite{b55}. Since our clinical concepts are derived from pre-annotated reports, potential bias is minimized. The list of clinical concepts associated with a pathology is well understood in the medical community and is straightforward for an expert to provide. The consultant radiologist in our study needed approximately 15 minutes to provide the list of concepts for lung cancer and healthy diagnoses.

This process resulted in a list of diagnostic phrases commonly used by experts in clinical workflows for detecting lung cancer in chest X-rays. To address sparsity and redundancies (e.g., "nodular opacity" vs. "nodular opacities"), we clustered these phrases under the guidance of the radiologist, creating consolidated clinical concepts for each label. These concepts, along with the original phrases used to define them, are summarized in Table~\ref{tab:concepts}. It is important to note that this task is required only once during the implementation of ClinicXAI and only when defining a new pathology for detection. Once the list of concepts is established, the framework operates automatically. By making these consolidated concepts publicly available, we enable other users to bypass this step, removing the need for additional expert input.

\begin{figure}[tb]
\centering
    \includegraphics[width=0.47\textwidth]{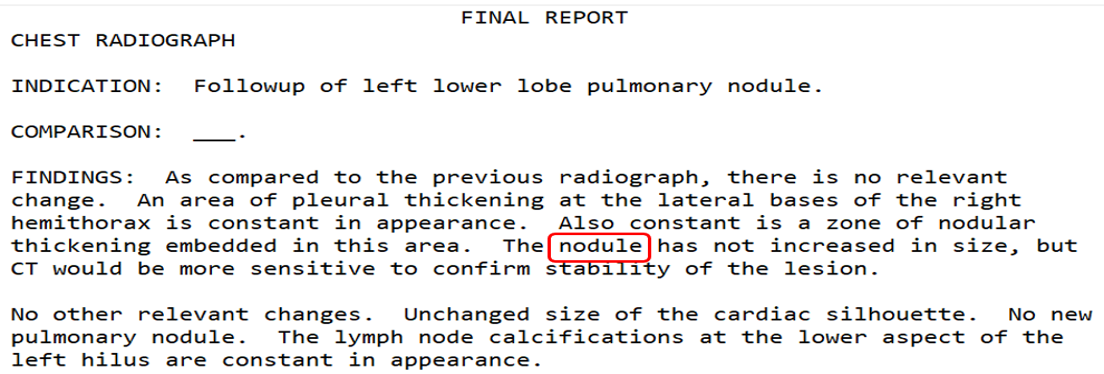}
\caption{Example of a cancerous radiology report from the MIMIC-CXR dataset. Clinical concepts extracted (`Nodule') are highlighted by bounding box. Note the negative mention in the final paragraph is not extracted.}
\label{fig:report}
\end{figure}

Each chest X-ray in the MIMIC-CXR dataset is linked to a free-text radiology report (Figure \ref{fig:report}), similar to clinical practice. To transform a radiological report into a concept vector representing the presence of the six resulting clinical concepts, we first preprocess the report to clean the text. We then identify which of the original phrases associated with each concept appear in the formatted text.

To address variability in report structures and following radiologist guidance, our analysis focuses exclusively on the FINDINGS and IMPRESSION sections of radiological reports, excluding sections such as HISTORY and COMPARISON, which may introduce noise. The preprocessing pipeline includes segmenting the report into individual sentences, removing punctuation and formatting characters (e.g., newlines), and converting all text to lowercase. Sentences containing fewer than two words or beginning with a negating statement (e.g., "there is no") are discarded. Additionally, sentences with terms likely to induce false positives (e.g., "nipple shadow" in the context of "nodular opacity is more likely to be a nipple shadow") are excluded. Finally, segments of sentences following negating phrases (e.g., "should not be mistaken for") are removed.

This process generates a refined set of formatted sentences that are free of negations and misleading terms, making them suitable for concept detection. The concept vector for each report is represented as a binary array, where each entry indicates the presence or absence of a specific clinical concept within the formatted sentences.

As shown in Table~\ref{tab:concepts}, clinically relevant phrases for each concept are classified into two categories: Type A phrases, which are fully encapsulated terms (e.g., "mass," "nodule"), and Type B phrases, which consist of flexible word collections that can appear in various orders (e.g., "hilus enlarged" as "the hilus appears enlarged" or "enlarged hilus"). For Type A phrases, the corresponding concept in the report's concept vector is set to 1 if the exact phrase occurs in at least one formatted sentence. For Type B phrases, each formatted sentence is analyzed to determine whether all words in the phrase, or their synonyms (e.g., "hilum" = "hilus" = "hilar"), are present. If every word or its synonym is identified within a sentence, the concept is set to 1 in the concept vector.

Following this process, each radiology report and corresponding chest X-ray is assigned a concept vector indicating the presence of six clinical concepts. As a final preprocessing step, the Unremarkable concept, which represents the Healthy class, is set to 0 if any other concept is detected. This adjustment prevents mislabeling cases where a report describes one lung as clear but mentions a pathology in the other, which could incorrectly assign the Healthy class.

\section{Evaluation of Existing XAI}

In this section, we evaluate the performance of widely used image-based and text-based XAI techniques. The image-based methods are assessed for their ability to identify medical ground truth pathology locations, represented as bounding boxes, and their level of agreement with one another. Text-based methods are evaluated based on their capacity to capture ground truth clinical concepts extracted from the radiology reports corresponding to each chest X-ray. Additionally, an expert radiologist reviewed and assessed the clinical utility of explanations generated by each approach for a selected subset of chest X-rays.

\subsection{Post-Hoc Image-Based XAI}\label{evalimagexai}

For the evaluation of post-hoc image-based XAI methods, we select both gradient- and perturbation-based techniques. Specifically, we use gradient-based Grad-CAM and perturbation-based LIME and SHAP, all of which are widely used and well-documented in the literature \cite{b56}. These methods are implemented using their respective Python libraries and applied post-hoc to an InceptionV3 model, which was trained using 10-fold cross-validation on our MIMIC-CXR dataset. InceptionV3 is an efficient, widely-used, and high-performing model architecture for chest X-ray pathology detection \cite{b57} \cite{b58} \cite{b59}, and achieves a mean F1 score of 0.782 on our MIMIC-CXR test set (Table~\ref{pred_performance}). All experiments are conducted using an NVIDIA GTX 1060 6GB GPU.

These image-based XAI techniques produce explanations in the form of heatmaps of pixel importance scores, where highly scored pixels are deemed most important to the classification decision of the model. In a clinically valid explanation, these high-scoring pixels should correspond to the clinically relevant regions of the image, such as a tumour in the context of lung cancer detection in chest X-rays. Despite the methodological differences among the techniques, we would also expect their explanations to highlight similar regions, provided they are accurately capturing the true medical signals within the image.

Based on this, we evaluate these explanations according to two criteria: 1) the agreement between the techniques and 2) their ability to reliably capture the medical ground truth.

\begin{figure}[tb]
\centering
\begin{subfigure}{0.45\textwidth}
    \includegraphics[width=\textwidth]{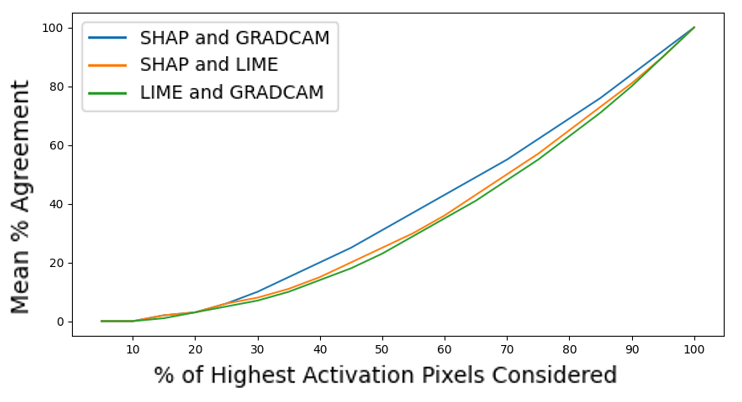}
\end{subfigure}
\begin{subfigure}{0.45\textwidth}
    \includegraphics[width=\textwidth]{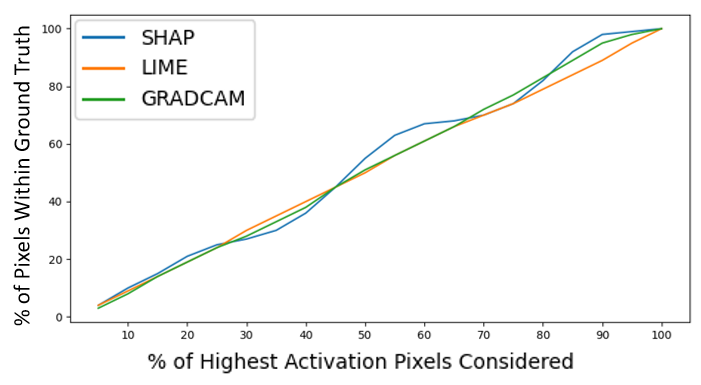}
\end{subfigure}
\caption{For LIME, SHAP and Grad-CAM, (a) shows mean pixel overlap between techniques on the MIMIC-CXR test set. (b) shows mean medical ground truth captured on the VinDr-CXR test set. Each technique is applied post-hoc to InceptionV3, trained with 10-fold cross-validation on the MIMIC-CXR dataset. Error bars are excluded due to negligible size.}
\label{fig:pixeloverlap}
\end{figure}

To assess the agreement between techniques, we analyze the similarity of the top $n$\% highest-activating pixels across our 1948 MIMIC-CXR test set images for various values of $n$. These pixels are identified as the most critical to the model's classification decision. If the explanations reliably capture the correct important image regions, we would expect a high degree of similarity between the highest-activating pixels for small values of $n$. However, as shown in Figure \ref{fig:pixeloverlap}(a), this is not the case: LIME, SHAP, and Grad-CAM demonstrate consistently low levels of agreement.

\begin{figure*}[tb]
\centering
\begin{subfigure}{0.195\textwidth}
    \includegraphics[width=\textwidth]{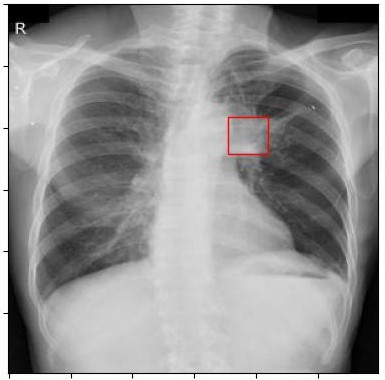}
    \caption{Ground Truth}
\end{subfigure}
\begin{subfigure}{0.195\textwidth}
    \includegraphics[width=\textwidth]{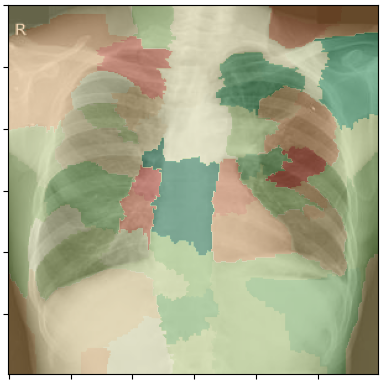}
    \caption{LIME}
\end{subfigure}
\begin{subfigure}{0.195\textwidth}
    \includegraphics[width=\textwidth]{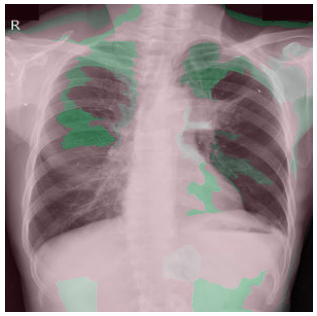}
    \caption{SHAP}
\end{subfigure}
\begin{subfigure}{0.193\textwidth}
    \includegraphics[width=\textwidth]{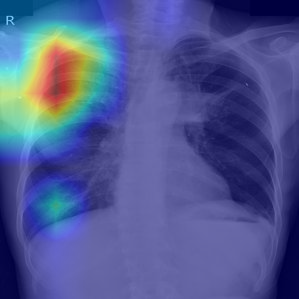}
    \caption{Grad-CAM}
\end{subfigure}
\begin{subfigure}{0.20\textwidth}
    \includegraphics[width=\textwidth]{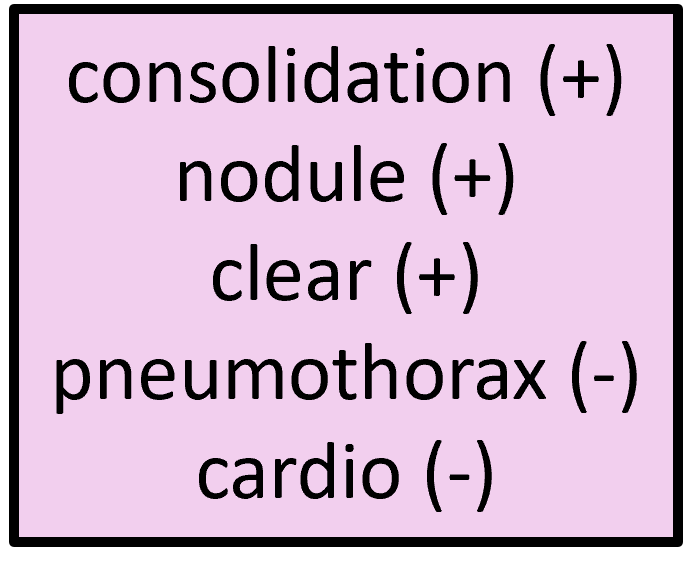}
    \caption{XCBs}
\end{subfigure}
\begin{subfigure}{0.42\textwidth}
    \includegraphics[width=\textwidth]{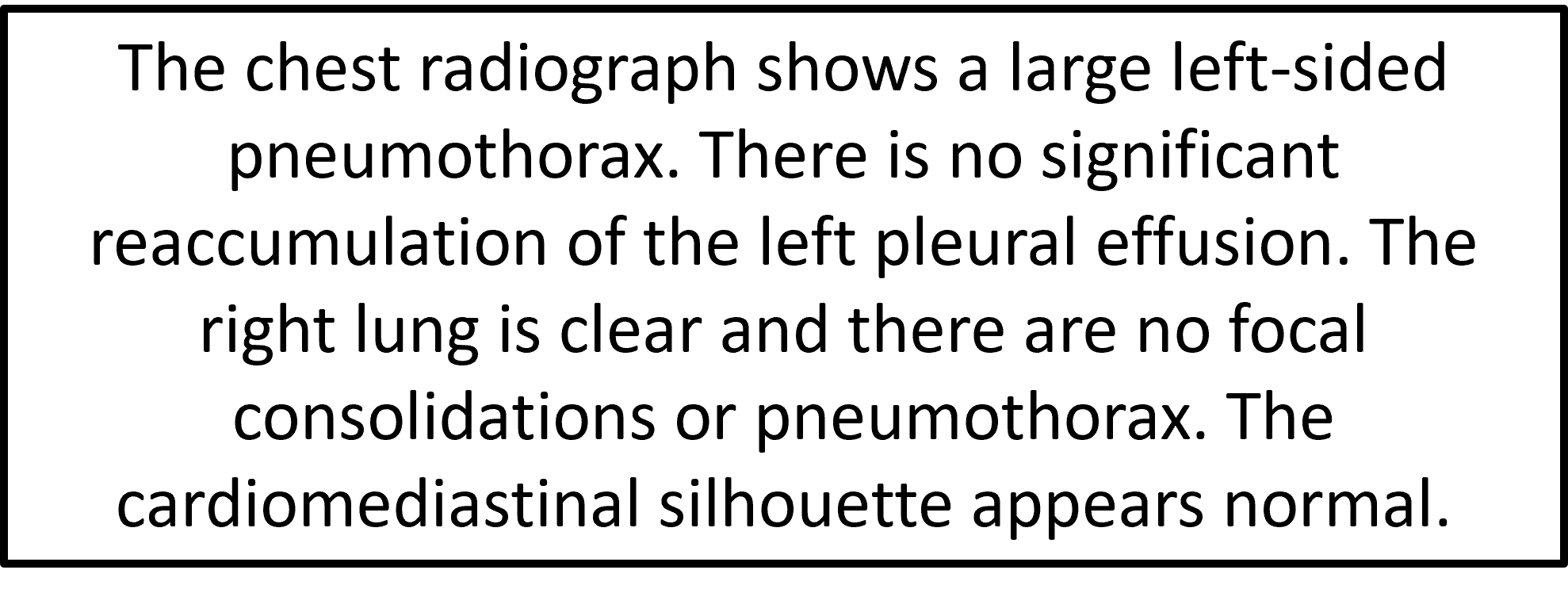}
    \caption{CXR-LLaVA}
\end{subfigure}
\begin{subfigure}{0.22\textwidth}
    \includegraphics[width=\textwidth]{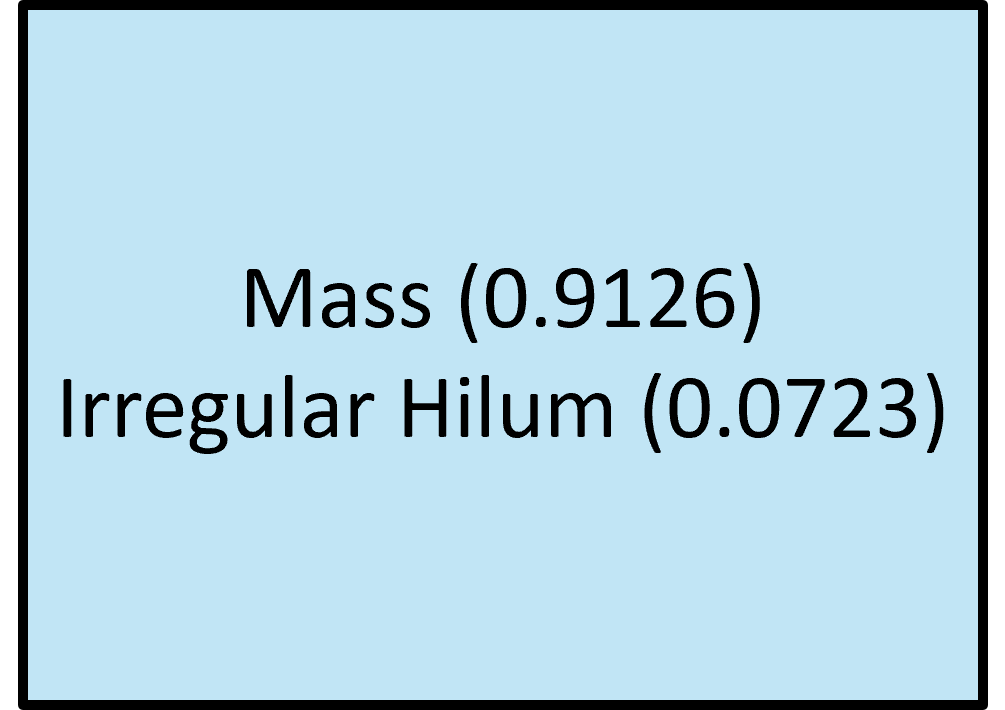}
    \caption{ClinicXAI}
\end{subfigure}
\caption{Explanations generated by each XAI technique for a cancerous chest X-ray. (a) shows the ground truth hilar mass. (b) shows LIME (most important = intense green). (c) shows SHAP (most important = green). (d) shows Grad-CAM (most important = red). (e) shows XCBs (concepts with the 5 highest absolute values, positive (+) or negative (-)). (f) shows the radiology report generated by CXR-LLaVA. (g) shows our expert-driven CBM, ClinicXAI, with the top 2 scoring concepts.}
\label{fig:bigexample}
\end{figure*}

We then compare the explanations generated by these techniques to the ground truth of a public dataset. Since MIMIC-CXR does not include pathology location annotations, we use the VinDr-CXR dataset for this evaluation. This dataset contains 826 cancerous chest X-rays with pathology location annotations in the form of bounding boxes, as provided by three independent radiologists. We evaluate the explanations generated by LIME, SHAP and Grad-CAM by calculating the proportion of the top $n$\% highest-activating pixels that fall within the ground truth bounding boxes across this dataset, for varying values of $n$. Clinically correct explanations should show high values of overlap with the ground truth for low values of $n$. As shown in Figure \ref{fig:pixeloverlap}(b), each technique consistently fails to capture the clinically relevant regions of the chest X-rays. An example of this behaviour is shown in Figure~\ref{fig:bigexample}.

\subsection{Text-Based XAI}\label{texteval}

\begin{table}[tb]
\small
\centering
\caption{
Proportions of ground truth clinical concepts captured by CXR-LLaVA, XCBs, and our approach ClinicXAI, evaluated on the 1948 image-report pair MIMIC-CXR test set.}
\begin{tabular}{|c|c|c|c|}
\hline
 & \multicolumn{3}{c|}{\textbf{Accuracy}} \\
\textbf{Model} & \textbf{Whole Test Set} & \textbf{Cancerous} & \textbf{Healthy} \\
\hline
ClinicXAI & \textbf{1.000} & \textbf{1.000} & \textbf{1.000} \\
CXR-LLaVA & 0.726 & 0.523 & 0.936 \\
XCBs & 0.852 & 0.801 & 0.893 \\
\hline
\end{tabular}
\label{concept_performance}
\end{table}

For the evaluation of text-based XAI techniques, we use CXR-LLaVA, a high-performing LLaVA-based multimodal large language model (LLM) specialized for chest X-ray pathologies, and XCBs, a recent CBM-based method that learns unsupervised concepts from chest X-rays and radiology reports using a cross-attention mechanism. These methods were selected due to their medical adaptations of frameworks which have been prevalent in recent literature—LLaVA models and Concept Bottleneck Models. Implementations of CXR-LLaVA and XCBs are publicly available in their respective studies.

Both of these methods generate textual explanations. CXR-LLaVA, when prompted with "Write a radiologic report on the given chest radiograph, including information about lung cancer" for a specific chest X-ray, produces a concise radiological report detailing any present pathologies or features. XCBs return a list of unsupervised concepts along with their corresponding confidence scores, which can be either positive or negative depending on the indication. We consider the five concepts with the highest absolute scores as the explanation, following the original study \cite{b52}. Example explanations can be found in Figure~\ref{fig:bigexample}.

We assess the clinical relevance of these explanations by evaluating their ability to reliably capture ground truth clinical concepts. We apply the concept extraction approach detailed in Section~\ref{conceptextraction} to all 1948 radiological reports in the MIMIC-CXR test set, and consider these extracted concepts to be the ground truth. Next, we apply the same concept extraction process to the report explanations generated by CXR-LLaVA and calculate the proportion of ground truth concepts captured in these explanations. Similarly, for XCBs, we evaluate the proportion of ground truth concepts captured by the top five absolute-scoring concepts in their generated explanations.

We found that CXR-LLaVA’s reports frequently produced false negatives for lung cancer diagnosis, instead emphasizing pneumothorax or cardiomegaly, likely due to imbalances in the training data \cite{b32}. On our MIMIC-CXR test set of 1948 chest X-rays and radiology reports, CXR-LLaVA achieved an overall accuracy of \textbf{72.6\%}, which dropped to \textbf{52.3\%} for cancerous scans (Table \ref{concept_performance}). CXR-LLaVA’s reports were competent at explaining diagnoses for chest X-rays with visually obvious large masses, but struggled to detect more subtle pathologies such as irregularities in the hilum or lung parenchyma. XCBs perform relatively well on our test set, capturing \textbf{85.2\%} of correct clinical concepts over all scans, dropping slightly to \textbf{80.1\%} when only considering cancerous scans. However, its explanations often included irrelevant and ambiguous concepts, such as "range" which could cause confusion.

\subsection{Radiologist Evaluation of XAI}\label{radiologisteval}

To assess the clinical relevance of the explanations generated by these approaches, we had a consultant radiologist with over 15 years of experience analyse the explanations generated for a random subset of 60 chest X-rays – 40 cancerous and 20 healthy. The number of images assessed is relatively low due to the time constraints of our expert, as explanations needed to be assessed for six XAI techniques (LIME, SHAP, Grad-CAM, CXR-LLaVA, XCBs and ClinicXAI) for each X-ray.

\begin{figure*}[tb]
\centering
\begin{subfigure}{0.42\textwidth}
    \includegraphics[width=\textwidth]{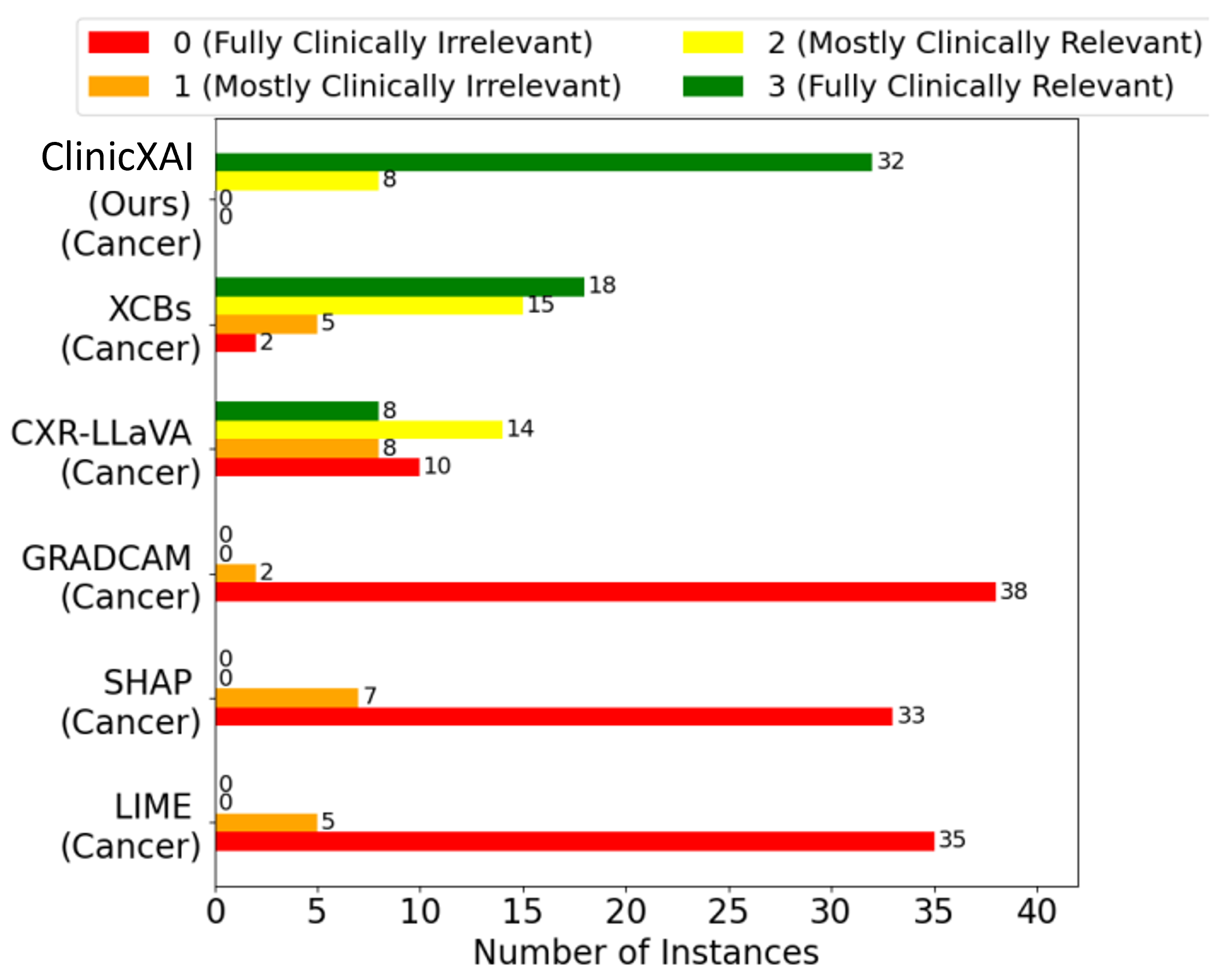}
    \caption{40 cancerous chest X-rays}
\end{subfigure}
\begin{subfigure}{0.42\textwidth}
    \includegraphics[width=\textwidth]{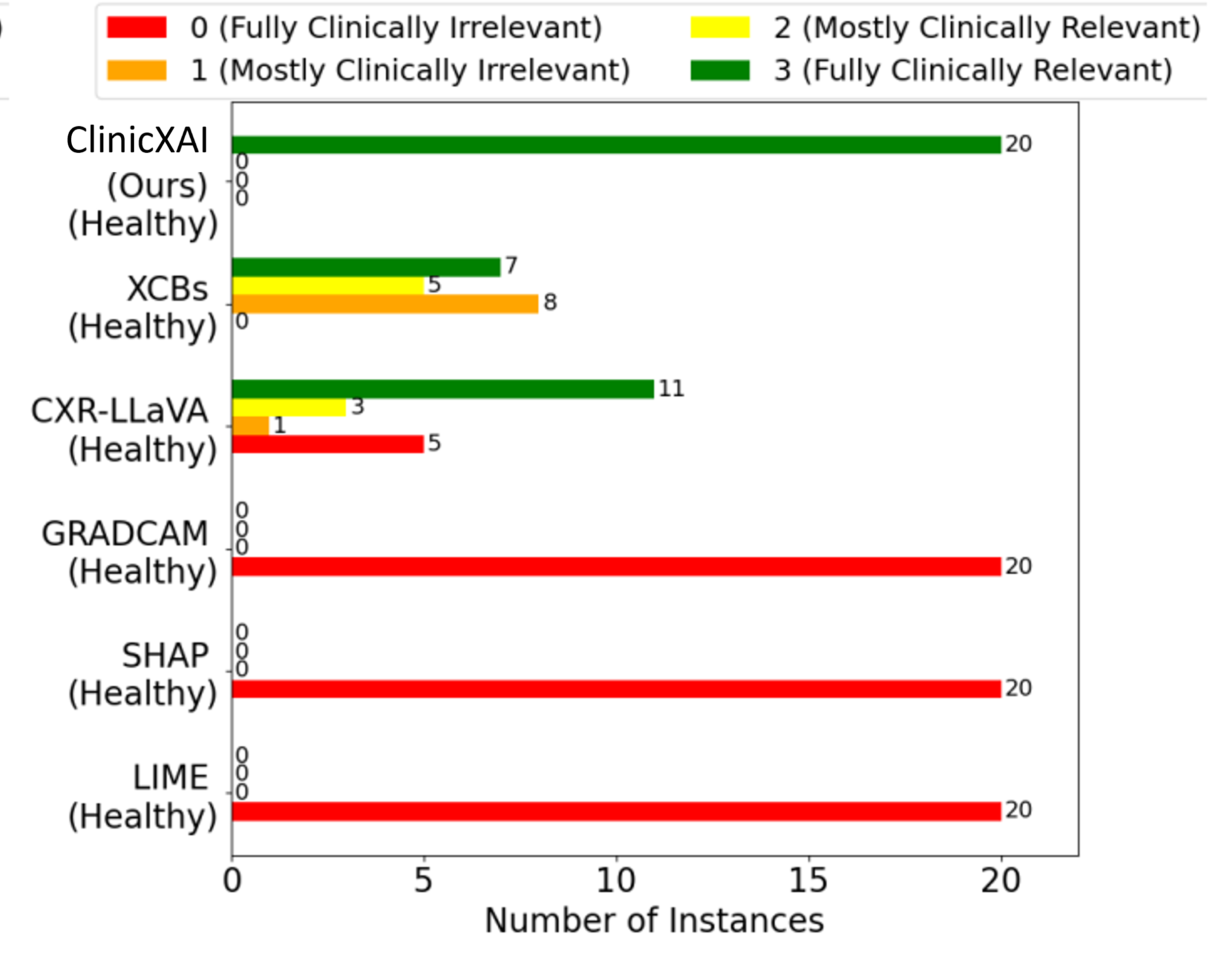}
    \caption{20 healthy chest X-rays}
\end{subfigure}
\caption{Analysis by an expert radiologist of explanations generated for a subset of 40 cancerous (a) and 20 healthy (b) chest X-rays for each of the XAI techniques evaluated in this study. The expert was asked to score explanations between 0 and 3 based on their clinical relevance (see legend).}
\label{fig:clinical}
\end{figure*}

We asked our expert to score each explanation between 0 and 3, where 0 indicates the explanation is fully clinically irrelevant, 1 indicates it is mostly clinically irrelevant, 2 indicates it is mostly clinically relevant, and 3 indicates that it is fully clinically relevant. We will refer to these scores as \textbf{Utility Scores}. Results are shown in Figure \ref{fig:clinical}.

LIME, SHAP and Grad-CAM score 0 on all healthy scans, as well as a strong majority of cancerous scans. The expert commented that each image-based XAI approach incorrectly highlighted areas of clear lung, and areas outside the lungs. This indicates a very low level of clinical relevance of explanations generated by these approaches. An example of this is shown in Figure \ref{fig:bigexample}(b-d). LIME, SHAP and Grad-CAM explanations for the cancerous X-ray in this figure scored 0, 1 and 0 respectively.

CXR-LLaVA has a much more variable performance, with a mean utility score of 2.0 (out of 3.0) on healthy scans and 1.5 on cancerous scans. The expert noted that CXR-LLaVA-generated reports frequently miss critical features like hilar masses or small nodules, while incorrectly identifying conditions such as scoliosis or pneumothorax. Some reports contained contradictory statements, such as indicating a lung was clear while also noting a mass. Figure \ref{fig:bigexample}(f) provides an example of a CXR-LLaVA report that received a score of 0, where the expert identified a false mention of pneumothorax and the omission of a hilar mass.

XCBs perform well on cancerous scans with a mean utility score of 2.225 (out of 3.0) - explanations are commonly scored as either mostly or fully clinically relevant. Performance on healthy scans is lower, with a mean utility score of 1.95. Our expert commented that XCBs were competent at identifying masses but also highlighted irrelevant concepts, such as ‘aortic’ on chest X-rays with no aortic abnormalities, or confusing concepts, such as ‘range’ or ‘contours’. It was also mentioned that XCBs incorrectly diagnosed effusions and pneumothorax on multiple occasions. An example is shown in Figure~\ref{fig:bigexample}(e). The XCBs explanation in this figure was given a score of 2 by the radiologist. Our expert stated that the mention of consolidation was false, but that the hilar mass was correctly identified. On healthy scans, our expert commented that XCBs had a high false positive rate, commonly identifying nonexistent abnormalities.

\section{ClinicXAI: Expert-Driven Concept-Based Explanations}

Our findings reveal that existing XAI techniques frequently underperform in specific medical contexts, such as lung cancer detection in chest X-rays, with explanations often failing to reliably capture clinically relevant features. To address this limitation and investigate the advantages of incorporating expert input into the design of clinically focused XAI methods, we propose ClinicXAI. This CBM-based approach diverges from unsupervised concept learning, as seen in XCBs, by leveraging clinical concepts curated by an expert radiologist. This ensures the explanations produced are both clinically relevant and optimized for utility in medical practice. This section introduces the ClinicXAI framework and evaluates its classification performance, interpretability, and robustness in comparison to other XAI techniques assessed in this study.

\begin{table*}[tb]
\small
\centering
\caption{Mean (and standard deviation) label prediction performance of ClinicXAI, compared to the InceptionV3 model and XCBs, on our MIMIC-CXR test set of 1948 chest X-rays using 10-fold cross-validation.}
\begin{tabular}{|c|c|c|c|c|c|c|}
\hline
\textbf{Model} & \textbf{Precision} & \textbf{Recall} & \textbf{F1} & \multicolumn{3}{c|}{\textbf{Accuracy}}\\
& & & & \textbf{Lung Cancer} & \textbf{Healthy} & \textbf{All} \\
\hline
ClinicXAI & \textbf{0.823 (0.03)} & \textbf{0.984 (0.01)} & \textbf{0.891 (0.02)} & \textbf{0.984 (0.01)} & \textbf{0.974 (0.01)} & \textbf{0.977 (0.01)}\\
InceptionV3 & 0.669 (0.02) & 0.960 (0.01) & 0.782 (0.00) & 0.960 (0.01) & 0.936 (0.00) & 0.939 (0.00)\\
XCBs & 0.697 (0.01) & 0.955 (0.00) & 0.801 (0.01) & 0.955 (0.01) & 0.947 (0.00) & 0.949 (0.01)\\
\hline
\end{tabular}
\label{pred_performance}
\end{table*}

\subsection{Framework Architecture}

\begin{figure*}[tb]
\centering
    \includegraphics[width=0.75\textwidth]{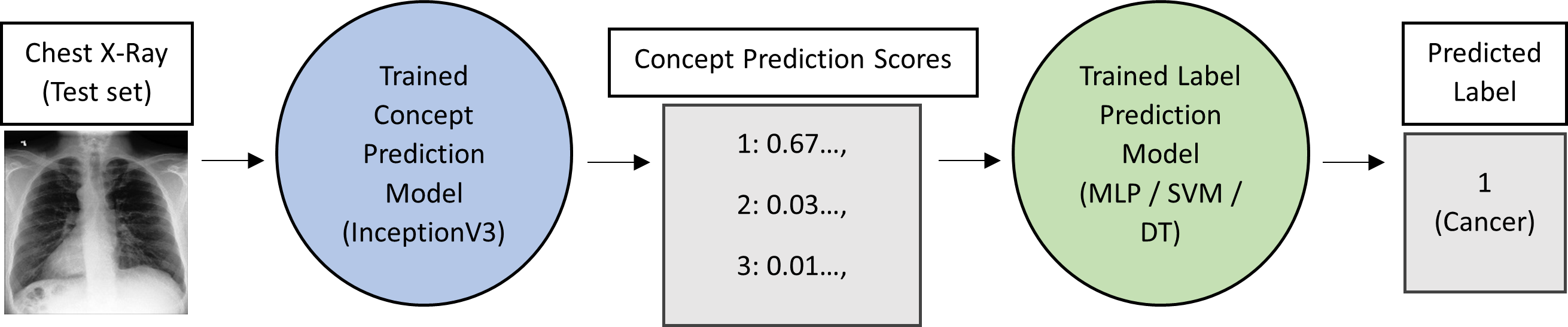}
    \caption{Inference pipeline for ClinicXAI takes a chest X-ray as input, which is fed into a trained concept prediction model, producing prediction scores for a pre-set list of clinical concepts. These scores are then input to a trained label prediction model, which outputs the binary classification label.}
    \label{fig:inference}
\end{figure*}

\begin{figure}[tb]
\centering
    \begin{subfigure}{0.4\textwidth}
        \includegraphics[width=\textwidth]{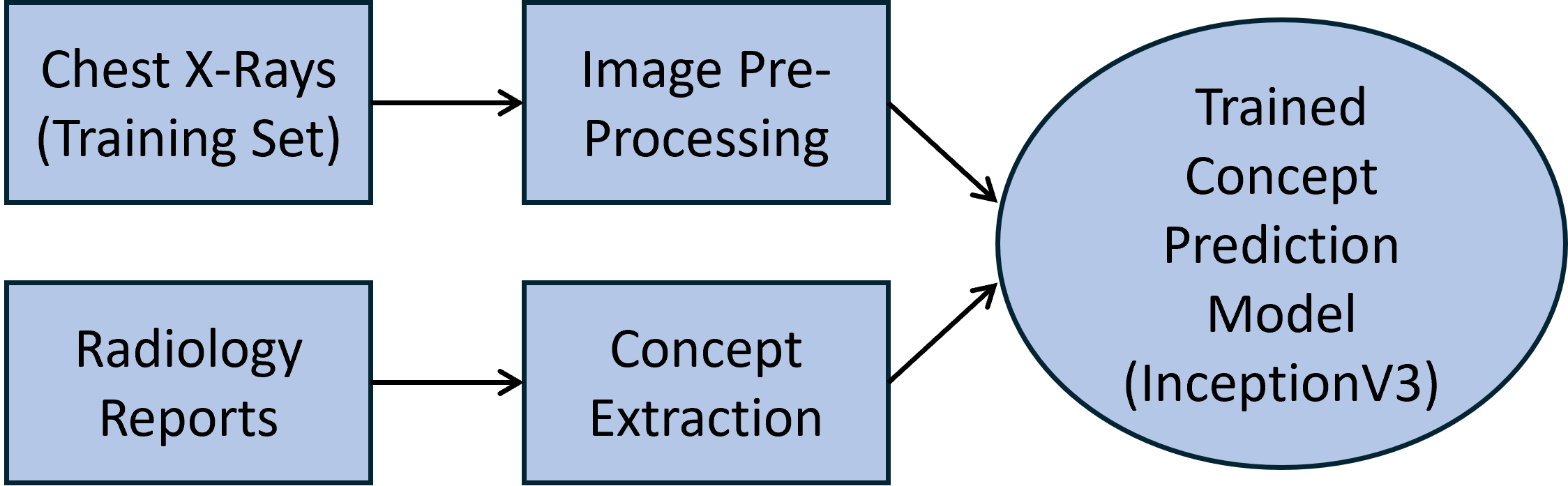}
        \caption{Concept Prediction Model}
    \end{subfigure}
    \begin{subfigure}{0.4\textwidth}
        \includegraphics[width=\textwidth]{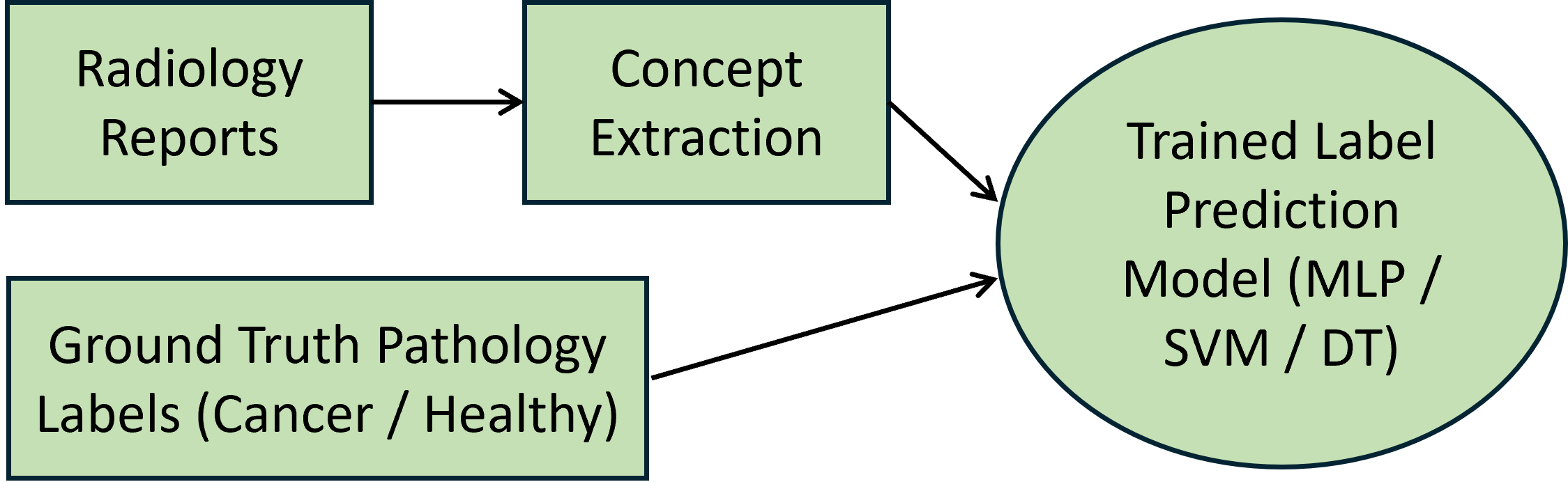}
        \caption{Label Prediction Model}
    \end{subfigure}
\caption{(a) shows the training pipeline for the concept prediction model taking as input Chest X-Rays and associated radiology reports. The presence of clinical concepts are extracted from the reports, and the resulting concept vectors are used to train the model to predict concept vectors for unseen chest X-rays. (b) shows the label prediction model, which takes radiology reports and ground truth pathology labels of their associated chest X-rays as input. Clinical concepts extracted from the reports are used to generate concept vectors, which, along with the pathology labels, are used to train the model to predict pathology labels from unseen concept vectors.}
\label{fig:training}
\end{figure}

As in the original CBM work \cite{b22}, we split the traditional classification pipeline into two separate models, as shown in Figures~\ref{fig:inference} and~\ref{fig:training}. We use the Independent CBM architecture as we are focusing on the clinical usability of explanations generated using the intermediate concept step, rather than the joint CBM architecture which merges the process into one end-to-end model.

The first model in the pipeline is the concept prediction model, which during inference takes a chest X-ray, and outputs a prediction score for each of a pre-determined list of concepts, chosen by a radiologist (Table~\ref{tab:concepts}). This model is trained using both chest X-ray images and associated radiology reports from the MIMIC-CXR training dataset. Images are pre-processed, and clinical concept vectors are acquired for each associated report, as detailed in Section~\ref{conceptextraction}. These image-vector pairs are used to train the model to predict the presence of clinical concepts for unseen chest X-rays. The output is a concept vector containing confidence scores, each representing the model's confidence in the presence of a specific clinical concept. As in the original CBM work we use an InceptionV3 architecture, trained with a batch size of 16 and learning rate of 0.0001.

The second model is the label prediction model, which during inference uses the concept prediction scores generated by the previous model to predict the image label (Lung Cancer or Healthy). This model is trained on the radiology reports and pathology labels from our training dataset. Clinical concept vectors are again generated for each report, and these vectors, paired with their corresponding pathology labels, are used to train the model. The model is then tasked with predicting pathology labels for unseen concept vectors generated by the concept prediction model. The original CBM work uses a Multilayer Perceptron (MLP) model architecture for label prediction. During the design of this framework, we evaluated several architectures, including Support Vector Machines (SVMs) and Decision Trees, to identify the most suitable approach. Decision Trees demonstrated superior performance on our dataset compared to SVMs and MLP models. While the framework is flexible and can accommodate various architectures, all results presented in this study are therefore based on the Decision Tree architecture.

As an explanation framework, ClinicXAI provides interpretability by outputting the two clinical concepts assigned the highest confidence scores by the concept prediction model for a given chest X-ray. An illustrative example of a ClinicXAI explanation is presented in Figure~\ref{fig:bigexample}.

\subsection{Pathology Classification Performance}

Table~\ref{pred_performance} presents the classification performance of ClinicXAI in comparison to state-of-the-art CBM-based XAI method XCBs and the standard InceptionV3 model. The InceptionV3 model serves as the basis for evaluating post-hoc image-based XAI earlier in this study. All models are trained using 10-fold cross-validation on our MIMIC-CXR dataset.

ClinicXAI demonstrates high classification performance, achieving a mean F1 score of \textbf{0.891} on the MIMIC-CXR test set, which consists of 1,948 chest X-rays. This performance surpasses that of both the standard InceptionV3 model and the XCB framework. ClinicXAI also exhibits higher classification accuracy across both target classes, further emphasizing its effectiveness in this clinical context.

\subsection{Interpretability and Clinical Utility}

Since ClinicXAI generates text-based explanations in the form of the two highest-confidence clinical concepts for a given chest X-ray, we evaluate its interpretability using the same methodology applied to other text-based approaches in this study, CXR-LLaVA and XCBs. Specifically, we assess its ability to reliably capture ground truth clinical concepts, following the evaluation procedure detailed in Section~\ref{texteval}.

The results, presented in Table~\ref{concept_performance}, indicate that ClinicXAI captures \textbf{100\%} of the ground truth clinical concepts in the MIMIC-CXR test set of 1,948 chest X-rays and associated reports. This performance surpasses that of XCBs (\textbf{80.1\%}) and CXR-LLaVA (\textbf{72.6\%}), with the most striking difference observed in ClinicXAI's ability to accurately identify lung cancer-specific concepts. These findings support our hypothesis that incorporating expert input when designing an explainable model framework enhances its interpretability. By ensuring that the concept space is clinically relevant, ClinicXAI aligns more closely with real-world clinical data, thereby increasing its interpretability and therefore usability to medical professionals.

To further assess the clinical utility of ClinicXAI, we had a consultant radiologist evaluate the clinical utility (as defined in Section~\ref{radiologisteval}) of its explanations for a subset of 60 chest X-rays. The results, comparing the clinical utility of our approach with all other XAI methods evaluated in this study, are presented in Figure~\ref{fig:clinical}. Explanations were assigned a score between 0 (indicating complete clinical irrelevance) and 3 (indicating complete clinical relevance).

ClinicXAI provided fully clinically relevant explanations for all 20 healthy chest X-rays, achieving a mean utility score of 3.0, the highest possible value. For the 40 cancerous chest X-rays, the mean utility score was 2.8, outperforming all other XAI methods assessed in this study. All explanations which did not receive the maximum score of 3 were instead given a score of 2, indicating that they were mostly clinically relevant. The radiologist attributed this to disagreements in the definitions of "mass" and "nodule."

\subsection{Robustness}

\begin{figure*}[tb]
\centering
    \begin{subfigure}{0.32\textwidth}
        \includegraphics[width=\textwidth]{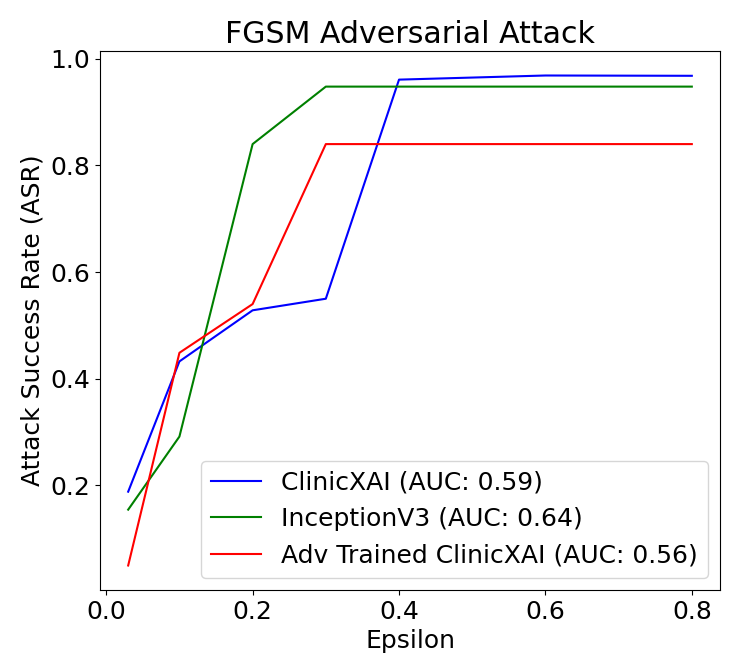}
        \caption{FGSM}
    \end{subfigure}
    \begin{subfigure}{0.32\textwidth}
        \includegraphics[width=\textwidth]{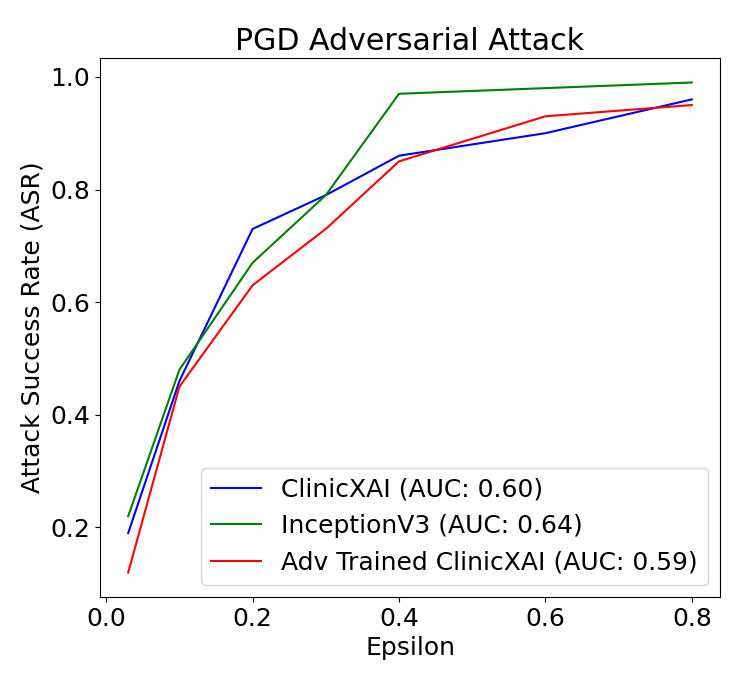}
        \caption{PGD}
    \end{subfigure}
    \begin{subfigure}{0.32\textwidth}
        \includegraphics[width=\textwidth]{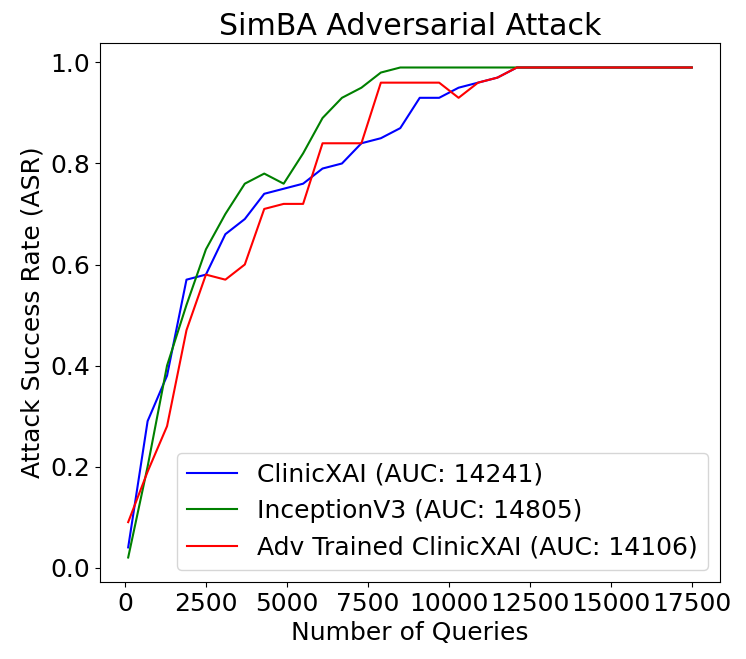}
        \caption{SimBA}
    \end{subfigure}
\caption{Attack Success Rate (ASR) curves for FGSM, PGD,
and SimBA. The legends display the
Area Under the Curve (AUC) values, where higher AUCs
correspond to steeper curves, indicating lower robustness of
the models to the attack.}
\label{fig:robust}
\end{figure*}

In addition to evaluating the interpretability and clinical utility of our expert-driven approach, we also assess its robustness. We assess the robustness of ClinicXAI by comparing it to the standard InceptionV3 model using three commonly used adversarial attack methods: SimBA (black-box), and FGSM and PGD (white-box) applied to our MIMIC-CXR test set. These attacks are typically evaluated using Attack Success Rate (ASR) curves, where ASR is computed over a range of parameter values for each attack. For white-box attacks like FGSM and PGD, the variable parameter is the epsilon value, while for the black-box SimBA, which iteratively queries the model to adjust adversarial perturbations, the variable is the number of queries.

ASR is defined as the proportion of adversarial examples that successfully lead to misclassification by the model, with a higher ASR indicating lower robustness to the attack. The ASR curves for each attack method are presented in Figure~\ref{fig:robust}, with the Area Under the Curve (AUC) values provided in the figure legends. A higher AUC corresponds to steeper curves, reflecting reduced robustness to the attack. ClinicXAI demonstrates superior robustness compared to InceptionV3 across all three adversarial attacks.

To further enhance the robustness of ClinicXAI, we implement adversarial training, a widely used defense strategy to strengthen classification models against adversarial attacks. For our adversarial training set, we generate adversarial images from the MIMIC-CXR validation set, consisting of 1948 chest X-rays, using all three adversarial attack methods. This results in an adversarial training set of 5844 images. We then fine-tune ClinicXAI on this dataset over 200 epochs, setting epsilon to 0.1 for both FGSM and PGD, as this is the threshold where adversarial images become human-distinguishable. As shown in Figure~\ref{fig:robust}, applying adversarial training to ClinicXAI further increases its robustness to these attacks.

\section{Discussion}

This study underscores the critical role of incorporating expert input into the development of interpretable and robust deep learning systems for medical diagnostics, particularly for lung cancer detection in chest X-rays. While recent advancements in explainable AI (XAI) techniques have shown promise in mitigating the opacity of black-box models by enhancing their transparency, this work highlights the substantial limitations of existing post-hoc and text-based XAI methods when applied in the medical domain. These methods frequently fail to generate clinically relevant or reliable explanations, hindering their adoption in high-stakes environments such as healthcare, where trust and precision are of paramount importance.

The evaluation of existing XAI techniques LIME, SHAP, Grad-CAM, CXR-LLaVA, and XCBs, revealed their limited alignment with clinical ground truth and expert assessments. Post-hoc image-based methods exhibited minimal agreement with one another and consistently failed to localize clinically significant regions, such as tumour sites. Although text-based approaches like CXR-LLaVA and XCBs demonstrated more potential, their utility was constrained by high false-negative rates, irrelevant concepts, and generation of ambiguous or contradictory explanations.

The proposed expert-driven concept bottleneck model, ClinicXAI, demonstrated significant improvements over these methods in clinical utility and interpretability, while maintaining superior classification accuracy and robustness against three widely utilized adversarial attacks. By integrating radiologist-defined clinical concepts during training, ClinicXAI generated explanations that consistently aligned with expert diagnostic criteria. Notably, ClinicXAI achieved 100\% accuracy in capturing ground truth clinical concepts, underscoring the efficacy of incorporating domain expertise into explainable model design. Furthermore, the clinical utility evaluation revealed that the explanations produced by ClinicXAI were highly relevant, as assessed by an experienced radiologist.

The integration of expert knowledge into the design of interpretable and robust deep learning models bridges the gap between technical performance and clinical relevance. The success of ClinicXAI suggests that domain-specific expertise is crucial for the development of reliable and clinically applicable AI systems. Future research should aim to extend the scope of expert-driven frameworks to encompass other medical imaging modalities and pathologies. 

While ClinicXAI is not presented as a groundbreaking innovation, its performance highlights the transformative potential of expert-driven explainable models in advancing AI-assisted medical diagnostics. Although the reliance on predefined clinical concepts may raise concerns about scalability to new pathologies, acquiring such concept lists is a straightforward, one-time task. To facilitate adoption, future iterations will include a multiclass implementation of the approach, along with publicly available clinical concept lists for new pathologies, eliminating the need for repeated concept-gathering efforts.


\end{document}